\documentclass[runningheads]{llncs}
\usepackage{graphicx}
\usepackage{amsmath,amssymb} 
\usepackage{color}
\usepackage[width=122mm,left=12mm,paperwidth=146mm,height=193mm,top=12mm,paperheight=217mm]{geometry}
\usepackage{ dsfont }

\DeclareMathOperator*{\argmin}{argmin}
\usepackage{algorithm}
\usepackage{algorithmic}
\usepackage{setspace}
\newcommand{\ve}[1]{\mathbf #1}
\usepackage{epsfig}
\usepackage{array}
\usepackage{float}
\usepackage{caption}
\usepackage{enumitem}
\begin{document}
\pagestyle{headings}
\mainmatter
\def\ECCV16SubNumber{1096}  
\title{Fast Semantic Image Segmentation with High Order Context and Guided Filtering} 

\titlerunning{Fast Semantic Image Segmentation}

\authorrunning{Shen \emph{et~al.}}

\author{Falong Shen, Gang Zeng}
\institute{Peking University}

\maketitle
\begin{abstract}
This paper describes a fast and accurate semantic image segmentation approach that encodes not only the discriminative features from deep neural networks, but also the high-order context compatibility among adjacent objects as well as low level image features.
We formulate the underlying problem as the conditional random field that embeds local feature extraction, clique potential construction, and guided filtering within the same framework, and provide an efficient coarse-to-fine solver. 
At the coarse level, we combine local feature representation and context interaction using a deep convolutional network, and directly learn the interaction from high order cliques with a message passing routine, avoiding time-consuming explicit graph inference for joint probability
distribution. 
At the fine level, we introduce a guided filtering
interpretation for the mean field algorithm, and achieve accurate object
boundaries with 100× faster than classic learning methods. 
The two parts are connected and jointly trained in an end-to-end fashion. Experimental results on Pascal VOC 2012 dataset have shown that the proposed
algorithm outperforms the state-of-the-art, and that it achieves the
rank 1 performance at the time of submission, both of which prove the
effectiveness of this unified framework for semantic image segmentation.
\end{abstract} 
\section{Introduction}
\begin{figure}
  \centering
  \includegraphics[width=0.9\columnwidth]{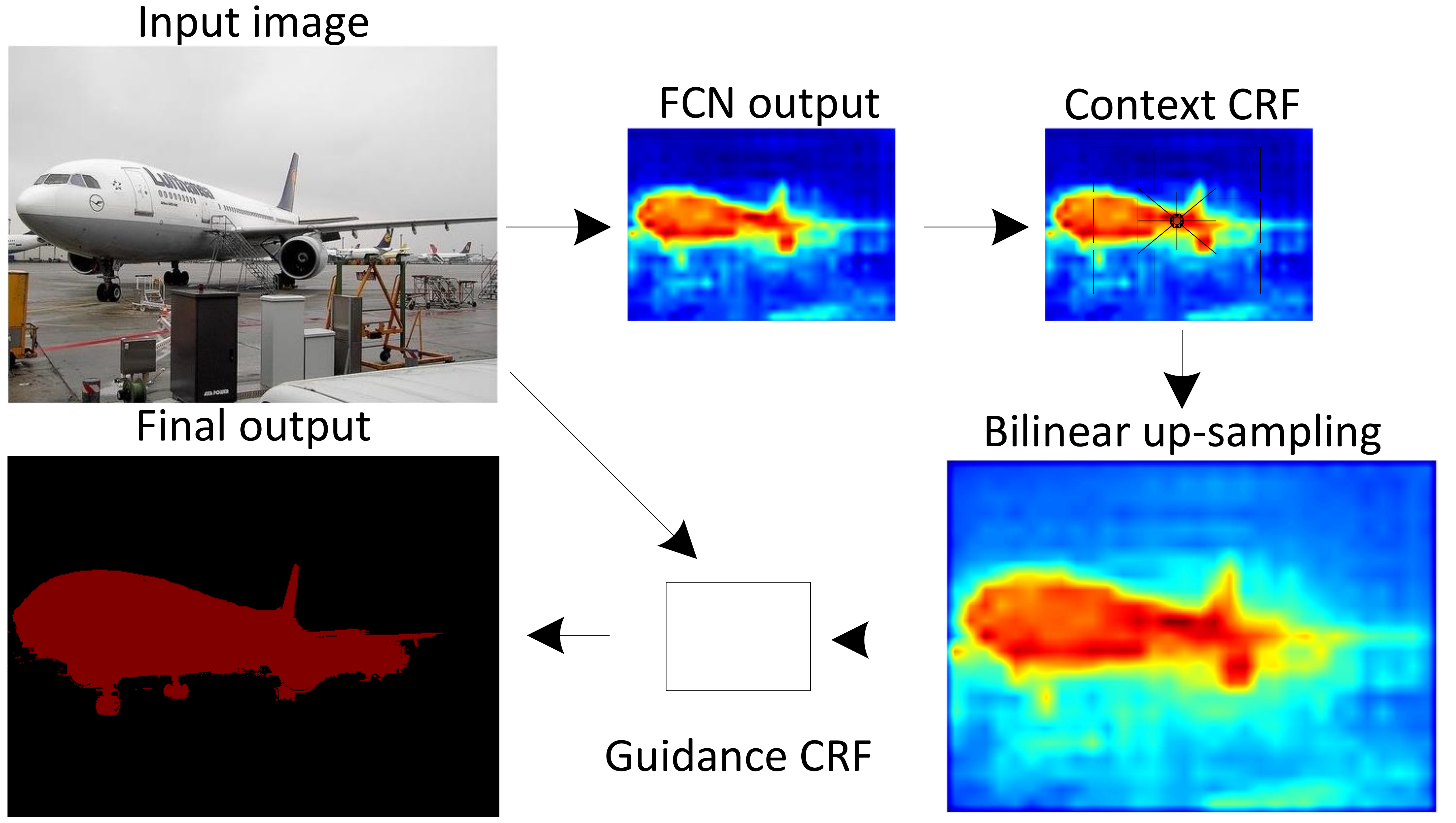}
  \caption{Schematic visualization of our model. At the coarse level, the context CRF is performed on the coarse segmentation score map to encode context information. At the fine level, the guidance CRF is adapted to delineate the object boundary to make it follow the edges in the input image.}
  \label{figure:illustration}
\end{figure}
The task of semantic image segmentation is to assign a label to each pixel in an image. Compared with image classification, semantic image segmentation provides a position-aware semantic understanding of the image through a structural prediction framework.
Recent advances in semantic image segmentation mainly rely on Fully Convolutional Networks (FCN)  and conditional random field (CRF).
During the past years, convolutional neural networks have made a series of breakthroughs on the task image classification \cite{krizhevsky2012imagenet,ioffe2015batch,he2015deep}. Deep networks naturally integrate multi-level hierarchical features and classifier by lots of stacked layers in an end-to-end fashion. FCN transfers the recognition network in image classification by fine-tuning to semantic image segmentation in order to harness the learned deep feature representation \cite{long2014fully}.

Different from image classification, the task of segmentation needs to determine object position, shape and boundary, which relies on local contents. Pooling layers in convolutional neural network tend to tolerate the object translation and deformation but decrease the ability for locating and separating objects from the neighboring context. Probability graphic model is natural to be used as it is a structural prediction task to assign pixel-wise labels. In particular, CRF has observed widespread success in semantic image segmentation \cite{russell2009associative,krahenbuhl2012efficient}. However, correctly optimizing CRF requires exact inference in the learning stage, which costs too much time \cite{zheng2015conditional,lin2015efficient}. Instead of explicit global probability representation in CRF, we propose to use a series of classifiers to encode interactions between each node. Our model resembles the error-correcting iterative decoding methods in \cite{ross2011learning,tu2008auto}. We propose an alternative view in the message passing stage in mean field algorithm and update the marginal distribution by collecting message from neighborhood regions. The message estimator is directly modeled with region features consisting of estimated labels and deep convolutional feature.

Designing a strong feature representation is the key challenge in semantic image segmentation. Supervised deep learning feature representation, estimated label map and low level image features are most often used feature for semantic image segmentation. Our contributions are mainly on the exploitations of context clues and low level image features, which are detailedly described in the Section \ref{section:context} and the Section \ref{section:guidance} respectively. Therefore, in the following paragraphs, we will review three kinds of commonly used features and related works.

\textbf{Local feature} plays the most important role to classify individual pixels in semantic image segmentation.
Recently, deep learning approaches such as FCN \cite{long2014fully} have made immense success in semantic image segmentation. The key insight is the strong learning capacity of extremely deep networks such as VGG16 \cite{simonyan2014very} and ResNet \cite{he2015deep} on large-scale training data such as ImageNet \cite{russakovsky2014imagenet}. Taking input of an image with arbitrary size, FCN usually produces a much coarser resolution feature map because of sub-sampling. However, these sub-sampling layers are necessary to keep computational efficient and features invariance. Therefore it is necessary to apply some kind of image filtering for a clear and sharper object boundary.

\textbf{Context clue} represents the spatial-relationship between category labels which is important in structural prediction task.
It has been noted that context clue or high-order information plays a vital role in object detection and semantic image segmentation. Context comes into a variety of forms. Through minimizing the Gibbs energy, CRF is widely adapted for harnessing the context clue to make structural prediction.
However, these models are quite limited due to the time cost of graph inference for the derivation of partition function in each update of gradient descent.
Recently lots of methods to compute the derivation of partition function of CRF in deep learning framework have been proposed.
For example, Chen \emph{et al.} \cite{chen2014learning} attempted to approximate the global distribution using the product of marginal distributions of all cliques, different from mean field algorithm which only uses the unary marginal distributions to approximate the global distribution. Traditionally, the derivation of partition function can also be computed by Gibbs sampling \cite{kirillov2015generic}.
However, even simplifying the global distribution with only unary marginal distributions, it is still not efficient for graph inference as too many iterations are needed for stochastic gradient descent learning in convolutional neural network.
What's more, Lin \emph{et al.} \cite{lin2015efficient} has shown that piece-wise training was able to achieve better performance and faster convergence than pseudo-likelihood training throughout their experiment. These things imply the difficulties of jointly training of FCN and CRF.
Besides, though recently some works have explored the effectiveness of using fixed-pattern high-order cliques \cite{arnab2015higher,vineet2014filter}, CRF usually is restricted into unary cliques and pair-wise cliques.

Compared to the traditional CRF approach for structural prediction, auto-context \cite{tu2008auto} encoded the joint statistics by a series of classifiers based on the label context. For each classifier, the output of the last classifier is used as feature. Auto-context made an attempt to recursively select and fuse context label for structural prediction.
Another probability of encoding context information is learning the messages based on feature context \cite{lin2015deeply,ross2011learning}. The kind of feature context methods model the message estimator between each pair by stacking unary features, which is more similar to traditional CRF as they both rely on pair-wise message passing.
Label context methods are natural to encode high-order clique potential. Pixels with strong local feature clues often achieve high probabilities for their label and they can pass these information to their correlated neighbors. Each pixel can update its estimated label based on local feature and neighborhood supports.
Hierarchical label context \cite{munoz2010stacked} adapted a hierarchical super-pixel representation for coarse-to-fine prediction.

\begin{figure}
  \centering
  \includegraphics[width=1\columnwidth]{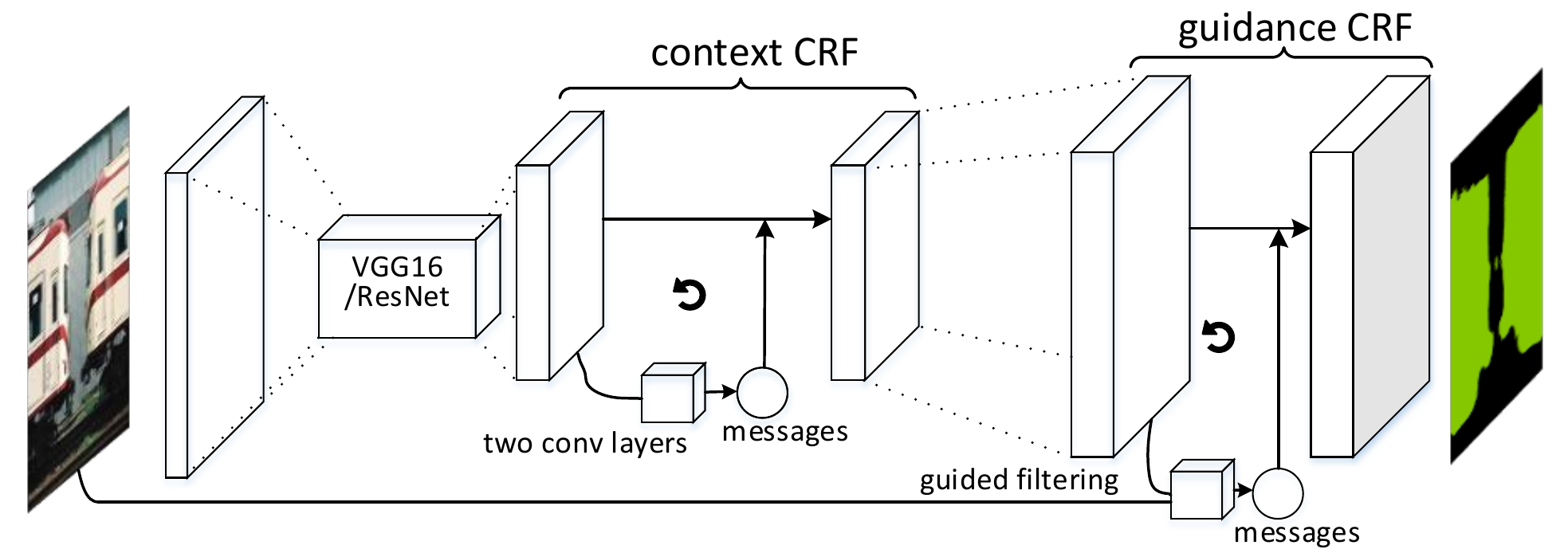}
  \caption{Model illustration. The coarse score map from FCN is  feed into the context CRF component, we use a convolutional block of two layers to model high order messages. The guidance CRF is then applied to refine the object boundary. The whole network is trained in an end-to-end fashion.}
  \label{figure:flowchart}
\end{figure}

\textbf{Low level feature} describes the low level image properties, such as image edges, texture and appearance homogeneity. Color histogram and gradient histogram are often used to obtain a clear and sharp boundary around objects.
Recently bilinear-filtering based CRF is popularly adapted for boundary localization. Combined with strong recognition capacity of convolutional neural network, bilateral filter based CRF has shown remarkable success in addressing the task of sharp boundary around object. Though the brute force implementation of bilateral filter is very slow, there are many speedy versions using the techniques of down-sampling \cite{adams2010fast} or quantization \cite{yang2009real}.
Besides, \cite{liu2015semantic} proposed a filter similar to the bilateral filter which can be processed on graphic process unit efficiently through the locally convolutional layer.
The guided filter is also an edge-preserving smoothing operator and it has better performance near edge. It can transfer the structure of guidance image to the filtering output, which is exactly what we want to do for the coarse segmentation map. What's more, the guided filter has a fast linear time algorithm, regardless of the kernel size. We plug the guided filter as the message passing step of pairwise CRF and called it guidance CRF. It leads to both fast process and high performance.

The main contributions of this paper have three folds.
\vspace{-2mm}
\begin{itemize}
    \item We propose a jointly trained model with high order context and guided filtering for semantic image  segmentation. The networks transfer the parameters from two contemporary classification models and are trained in an end-to-end fashion. It reaches an IOU score of $78.1\%$ and sets the new state-of-the-art on the Pascal VOC 2012 test set.
    \item We provide a new method to optimize CRF by encoding context information to update local estimation and introducing global nodes to make the structural prediction global consistency, which we called it context CRF. Experiments have verified the effectiveness of each component of our proposed model. Our proposed context CRF costs little time while bringing large performance gains.
    \item We plug in the guided filtering as a message passing step of guidance CRF and make the inference process for accurate boundary $100\times$ faster comparing to traditional bilateral filtering based fully connected CRF.
\end{itemize}
\clearpage
\section{Framework}
Let $I \in \ve I$ denote one input image and $\ve x \in \mathcal{X}$ is its ground truth segmentation label assignment in the dataset. Each pixel $i$ in the label assignment $\ve x = \{x_i,i=1,...,N\}$ takes a value from a pre-defined label set $\mathcal{L}=\{1,...,L\}$. Every label assignment $\ve x$ for image $I$ has a graph $G$ associated with it, where all the pixels forms the vertex set $\mathcal{V}$.

The conditional likelihood function for the image $I$ is
\begin{equation}
 P(\ve x|I;\theta)=\frac{1}{Z(I;\theta)}\exp[-E(\ve x,I;\theta)],
\label{equation:P}
\end{equation}
where $E(\ve x,I;\theta)$ is the Gibbs energy function with parameter $\theta$. $Z(I;\theta)$ is the partition function conditioned on the image $I$ and the model parameters $\theta$, $Z(I;\theta)=\sum_{\ve x}\exp[-E(\ve x,I;\theta)].$

The energy function in our formulation is written as
\begin{equation}
    E(\ve x, I;\theta) = E_{local}(\ve x,I;\theta) + E_{context}(\ve x;\theta) + E_{edge}(\ve x,I;\theta),
\end{equation}
where $E_{local}(\ve x,I;\theta)$ denotes the segmentation score map of FCN based on deep local feature, $E_{context}(\ve x;\theta)$ encodes the context clue to make structure prediction and $E_{edge}(\ve x,I;\theta)$ is designed to force the segmentation score map to follow the edges in the image.

The coarse segmentation score map of FCN has a lower resolution than original input image. We propose that to encode context information it is unnecessary to make predictions at original resolution. Therefore we have two steps as it is shown in the Figure \ref{figure:flowchart}. Firstly at the coarse level, we take local potential and context potential into consideration
\begin{equation}
    E_u(\ve x,I;\theta) = E_{local}(\ve x,I;\theta) + E_{context}(\ve x;\theta) .
\end{equation}
Note that we need to decouple each $x_i$ in this step, e.g., marginal potential with regard to each $x_i$. We solve it in the context CRF component.

After getting the marginal potentials, we up-sampling the segmentation score map to the same size as input image. Secondly at the fine level,  the total energy function is
\begin{equation}
    E(\ve x,I;\theta) = E_u(\ve x,I;\theta) + E_{edge}(\ve x,I;\theta) .
    \label{equation:Euedge}
\end{equation}
where $E_u(\ve x,I;\theta)$ can be treated as unary term as it has been expressed as summation of marginal potentials. Combined with edge potential, we can refine the segmentation score map and get a more accurate object boundary. It is solved in the guidance CRF component.

\subsection{Context Conditional Random Field at Coarse Level}
\label{section:context}

For a given image $I$, the FCN output is a segmentation score map and each pixel $i$ with label assignment $x_i$ has a unary potential $\phi_i(x_i,I_i;\theta)$ associated with it.
In our formulation, as done in \cite{vineet2014filter} and \cite{arnab2015higher}, we introduce $L$ hidden variables $\{y_l,l=1,...,L\}$ to describe the existence of categories in the image. Each hidden variable $y_i$ takes value from $\{0,1\}$ where $y_i=0$ represents that the $i$-th category appears in the image, otherwise the not.
The Gibbs energy of the label assignment $\ve x \in \mathcal{L}^N$ and $\ve y \in 2^L$ is
\begin{equation}
\begin{split}
    &E_u(\ve x,\ve y,I;\theta)= \\
    & \underbrace{\sum\limits_{i}\phi_i(x_i,I_i;\theta)+\sum\limits_l\phi_l(y_l,I;\theta)}_{local}+\underbrace{\sum\limits_{c}\psi_c(\ve x_c;\theta)+\sum_l\sum\limits_{i}{\psi_g(y_l,x_i;\theta)}}_{context},
\end{split}
  \label{equation:E}
\end{equation}
where $\phi_i$ and $\phi_l$  are the singleton node potentials. $\phi_i(x_i,I_i;\theta)$ is the potential of assigning $x_i$ to pixel $i$ based on local appearance descriptor extracted from $I_i$. $\phi_l(y_l,I;\theta)$ describes the existence of the $l$-th category in the image based on global image descriptor extracted from the whole image $I$. $\psi_c$ is defined on the high order clique $c$. $\psi_g$ is designed for the global consistency between global prediction $\{y_l\}$ and pixel-wise label assignment $\ve x$. $i$ indexes the pixel position in the image. The two context terms are independent of image $I$.

Our goal is to estimate the marginal potentials to approximate $ E_u(\ve x,\ve y,I;\theta)$, which is
\begin{equation}
    E_u(\ve x,\ve y,I;\theta) \approx \sum\limits_{i}\phi^u_i(x_i,I_i;\theta)+ \sum\limits_l\phi^u_l(y_l,I;\theta),
\end{equation}
Following the similar derivations of mean field algorithm \cite{koller2009probabilistic}, we can get proposed solution which is shown in Algorithm \ref{algorithm:marginal}.

\vspace{-5mm}
\begin{algorithm}
\caption{Marginal potential}
\textbf{input:} FCN unary potential $\phi_i(x_i,I_i;\theta)$ and $\phi_l(y_l,I;\theta)$, maximum iterations K. \\
\textbf{initialize}: $\phi^u_i(x_i|I;\theta)=\phi_i(x_i,I_i;\theta)$, $\phi^u_l(y_l|I;\theta)=\phi_l(y_l,I;\theta)$, k = 0.\\
\textbf{while not converge and $k < K$}\\
\vspace{-6mm}
\begin{enumerate}
\begin{spacing}{1.2}
  \item $\hat p(x_i|I;\theta) =\frac{1}{Z_i} \exp[-\phi_i^u(x_i;\theta)].$                     \hfill  $\triangleright$ Softmax
  \item $\hat p(y_l|I;\theta) =\frac{1}{Z_l}\exp[-\phi^u_l(y_l;\theta)].$
  \item  $\phi_i^u(x_i|I;\theta)=\phi_i(x_i,I_i;\theta)-\sum_c\mathds{E}_{\hat p(\ve x_{c \backslash i})}[\psi_c(\ve x_{c\backslash i},x_i;\theta)]\\- \sum_i\sum_l\mathds{E}_{\hat p(y_l)}[{\psi_g(y_l,x_i;\theta)}].$                                                         \hfill  $\triangleright$ Message passing
  \item $\phi^u_l(y_l|I;\theta)=\phi_l(y_l,I;\theta)-\sum_i\mathds{E}_{\hat p(x_i)}[{\psi_g(y_l,x_i;\theta)}].$
  \item k = k + 1.
\end{spacing}
\end{enumerate}
\vspace{-6mm}
\textbf{end while}\\
\textbf{output:} marginal potential $\phi_i^u(x_i;\theta)$ and $\phi^u_l(y_l;\theta)$.
\label{algorithm:marginal}
\end{algorithm}

 In Algorithm \ref{algorithm:marginal}, $\mathds{E}_{\hat p(x_i)}[{\psi_g(y_l,x_i;\theta)}]$ is the expectation of $\psi_g(y_l,x_i;\theta)$ over the estimated distribution of  $\hat p(x_i)$  and $\mathds{E}_{\hat p(y_l)}[{\psi_g(y_l,x_i;\theta)}]$ is the expectation of $\psi_g(y_l,x_i;\theta)$ over the estimated distribution of  $\hat p(y_l)$. The two terms can be treated as messages  reflecting the mutual interactions between the local label prediction $x_i$ and the global label prediction $y_l$.  $\mathds{E}_{\hat p(\ve x_{c \backslash i})}[\psi_c(\ve x_{c\backslash i},x_i;\theta)]$  is the expectation of $\psi_c(\ve x_{c\backslash i},x_i;\theta)$ over the estimated distribution of $\hat p(\ve x_{c \backslash i})$, which is  about the message passed from the high order clique $c$ to the local node $i$. We will show how to compute these three messages in the convolutional neural networks  in the following paragraphs.

  \textbf{$\bullet$ Two messages between local and global nodes}. It is straightforward to get the closed form expressions by the definition of expectation
\begin{equation}
\left \{
\begin{split}
\mathds{E}_{p(\hat y)}[{\psi_g(y_l,x_i;\theta)}]               &= \sum_{y_l}\hat p(y_l) \mu(y_l,x_i), \\
\mathds{E}_{\hat p(x_i)}[{\psi_g(y_l,x_i;\theta)}] &= \sum\limits_{x_i}\hat p(x_i) \mu(y_l,x_i).
\end{split}
\right.
\end{equation}
 Here we define $\psi_g(y_l,x_i;\theta)= \mu(y_l,x_i)$ and initialize $\mu(y_l,x_i)=\mathds{1}[x_i=l \wedge y_l=1]$, which encourages $y_l$ and $x_i$ to take consistent label. $\mu$ can be learned in the jointly training framework.

\textbf{$\bullet$ Message from clique to node.} It is a $L$-dimensional vector encoding the information of label distribution, which is difficult to get a analytical solution. Lin \emph{et al.} \cite{lin2015efficient} has tried to learn potential functions for each two-nodes clique but the inference is much slower and it costs lots of memory to save these joint potentials, e.g., it requires $L^2$ outputs for each pair-wise clique and for a $N$-nodes graph, there are up to $N^2$ pair-wise cliques. It is even much more difficult to learn a potential function for high order clique $c$  with more than two nodes. However, the high order clique is important to make use of the context information and learn the object shapes.

Instead of calculating the marginalization with regard to $\hat p(\ve x_{c\backslash i})$, we propose to construct the convolutional neural networks and directly learn the messages. We place two convolutional layers on the estimated probability map $\hat p(\ve x_c)$ in each iteration to capture the high order pattern
\begin{equation}
  \mathds{E}_{\hat p(\ve x_{c \backslash i})}[\psi_c(\ve x_{c\backslash i},x_i;\theta)] = U[\hat p(\ve x_{c}),x_i;\theta],
  \label{equation:highorder}
\end{equation}
where $U[\hat p(\ve x_{c}),x_i;\theta]$ is a scalar describing the compatibility of $x_i$ in the high order clique assignment $\ve x_c$. It can also be treated as a new classifier purely based on the estimated probability map, which is independent of image feature. As context information can come from objects far away, we set the size of high order clique very large, almost about half the image size.

Similar ideas can be found in the auto-context model \cite{tu2008auto}. They use a series of classifiers to update the estimated probability label map. In each iteration, the classifier is trained both on local image feature and estimated label context output by the previous classifier. However, in that work the classifiers in each iteration are piece-wise trained with the hand-crafted image features. Unlike their approach, we jointly train the classifier as well as feature layers in convolutional networks. Besides, the classifier in our approach is designed to model the message passed from high order clique $c$ to the node $i$, therefore it is only based on label context and independent of local image feature.

\subsection{Guidance Conditional Random Field at Fine Level}
\label{section:guidance}

The FCN provides a strong feature representation and we have encoded the context information to make structural predictions in the previous section. However, during to the employment of max-pooling layers and sub-sampling operations, the output of FCN is at much lower resolution and is coarse segmentation map. In previous works, the fully connected CRF with low level image features, e.g., color, coordinate, has been successfully used to enhance the object localization accuracy.

The guided filtering is an edge-preserving technique with nice visual quality and fast speed \cite{he2010guided}. We proposed to combine pair-wise CRF with guided filtering and jointly tune the whole networks to learn to align the segmentation results with the object boundary.

The guided filtering in our guidance CRF takes two inputs: (1) the coarse segmentation score map $\phi^u$ to be filtered and (2) the original color image $I$. The filtering result is
\begin{equation}
  g(x_i)=\sum\limits_i{w_{ij}(I)\phi^u_j(x_j)}.
\end{equation}
The weight $w_{ij}$ depends on the input color image $I$, which is used as the guidance image. Following the similar derivations in \cite{he2010guided}, the expression of $w_{ij}$ is
\begin{equation}
  w_{ij}=\frac{1}{|\omega|^2}\sum\limits_{k\in \omega_i,k\in \omega_j}{\bigg (1+ (\Sigma_k+\epsilon U)^{-1}\sum\limits_{c=1}^3(I^c_i-\mu^c_i)(I^c_j-\mu^c_j)) \bigg )}
  \label{equation:w}
\end{equation}
where $\mu_k$ and $\Sigma_k$ is the mean and $3 \times 3$ covariance matrix of image $I$ in window $\omega_k$, $U$ is $3 \times 3$ identity matrix and $|\omega|$ is the number of pixels in $\omega_k$. $\epsilon$ is a regularized parameter and we set it to 1 throughout our experiments.

Now we will introduce how to combine the pair-wise CRF with guided filtering. In the pairwise CRF model according to the Equation \ref{equation:Euedge}, the energy of a label assignment $\ve x$ is given by
\begin{equation}
  E(\ve x)=\underbrace{\sum_i\phi^u_i(x_i)}_{unary}+\underbrace{\sum_{i<j}\psi_p(x_i,x_j,I_i,I_j)}_{edge}.
\end{equation}
where the unary potential $\phi^u$ is the output of context CRF. Note that we have dropped the potentials for hidden variables $\ve y$ as it is not measured in our experiment. The pairwise potential $\psi_p$ in the fully connected CRFs has the form

\begin{equation}
  \psi_p(x_i,x_j,I_i,I_j)=\mu(x_i,x_j)k(I_i,I_j)
\end{equation}
where $\mu$ is the label compatibility function and the kernel $k(I_i,I_j)=w_{ij}$ as defined in the Equation \ref{equation:w}. $\mu$ is initialized by Potts model and it is jointly learned during training the whole networks to take interactions between labels into account. A mean-field algorithm is used to approximate the maximum a posterior solution as shown in Algorithm \ref{algorithm:guided}.

\begin{algorithm}[H]
\caption{\small Guidance CRF - Training}

\textbf{Forward}

\textbf{input:} Guiding image $I$, segmentation score map $\phi^u$, compatibility matrix $\mu$, weight parameter $\lambda$
\vspace{-3mm}
\begin{enumerate}
\begin{spacing}{1.2}
  \item  $q(x_i)=\frac{1}{Z_i}\exp[-\phi^u_i(x_i)]$. \hfill  $\triangleright$ Softmax

  \item   $g(x_i)=\sum_j{w_{ij}(I)q(x_j)}$   \hfill  $\triangleright$                Guided filtering

  \item  $m(x_i)=\sum_{x_i}\mu(x_i,x_j)g(x_j)$  \hfill  $\triangleright$              Compatibility transform

  \item  $\phi_i(x_i)=\phi^u_i(x_i) - \lambda m(x_i)$       \hfill  $\triangleright$             Local update
\end{spacing}
\end{enumerate}
\vspace{-8mm}
\textbf{output:} marginal potential $\phi$

\vspace{-3mm}
\rule{1\textwidth}{0.1mm}
\vspace{-5mm}

\textbf{Backward}

\textbf{input:} Guidance image $I$, segmentation score map $\phi^u$, compatibility matrix $\mu$, gradient of marginal potential $\frac{\partial L}{\partial \phi}$, weight parameter $\lambda$
\vspace{-3mm}
\begin{enumerate}
\begin{spacing}{1.5}
  \item $\frac{\partial L}{\partial  \phi^u_i}(x_i) = \frac{\partial L}{\partial  \phi_i}(x_i)$ ,$\frac{\partial L}{\partial  m}(x_i) = -\lambda \frac{\partial L}{\partial \phi_i}(x_i)$

  \item $\frac{\partial L}{\partial  \mu}(l_1,l_2) = \frac{\partial L}{\partial  m}(x_i)  g(x_j)$, $\frac{\partial L}{\partial  g}(x_i) = \frac{\partial L}{\partial  m}(x_j) \mu(x_i,x_j) $

  \item $\frac{\partial L}{\partial q}(x_i) =\sum_j w_{ij}(I) \frac{\partial L}{\partial g}(x_j)$

  \item $\frac{\partial L}{\partial  \phi_i}(x_i) = \frac{\partial L}{\partial  \phi_i}(x_i)$ +$\frac{\partial L}{\partial q}\frac{\partial q}{\partial \phi_i}(x_i)$
\end{spacing}
\end{enumerate}
\vspace{-8mm}
\textbf{output:}$\frac{\partial L}{\partial \phi^u}$, $\frac{\partial L}{\partial \mu}$
\label{algorithm:guided}
\end{algorithm}
\vspace{-5mm}

The forward pass in the training stage performs a softmax, a message passing step, a compatibility transform and a local update. As it is shown in the Froward part in Algorithm \ref{algorithm:guided}, all of these steps can be described by CNN layers. The parameters of guided filter depend on the spatial and appearance of the original image. Instead of directly computed by convolutional layers, the message passing step can be executed as one guided filtering, which can be computed very efficiently.

To back-propagate the segmentation error differentials w.r.t its input and network parameters in each layer, we have shown it in the Backward part in Algorithm \ref{algorithm:guided}. It is straightforward to perform back-propagation algorithm through the local update layer, the compatibility transform layer and the softmax layer.
For the message passing layer, the gradient w.r.t its input is
\begin{equation}
\frac{\partial L}{\partial g}(x_i) =\sum_j w_{ij}(I) \frac{\partial L}{\partial q}(x_j),
\end{equation}
which can also be calculated by performing guided filtering on the error differential map $\frac{\partial L}{\partial q}(x_j)$.

In the inference stage, as shown in \cite{he2015fast}, we down-sample (bilinear) the guidance image and score map, get the guidance parameters in the low resolution, up-sample (bilinear) the guidance parameter and get the filtering result. This operation accelerates this layer more than by $10\times$. We run three iterations in the inference stage.

With the marginal potentials, it is straightforward to get the marginal distribution $\hat p(x_i)=\frac{1}{Z_i}\phi_i(x_i)$. Given a training set $\{(I,\ve x),I\in \ve I,\ve x \in \mathcal{X}\}$, the target of CRF optimization is to learn the parameters $\theta^*$ to maximize the posterior probability of the training data,
\begin{equation}
  \theta^* = \argmin_{\theta}\sum_I\sum_i \log \hat p(x_i|I;\theta)+\frac{\lambda}{2}||\theta||^2_2
\end{equation}
Here $I$ is the training image and $x_i$ is the ground truth segmentation label for pixel $i$ in this image; $\lambda$ is the weight decay parameter. The program can be optimized by the standard stochastic gradient descent solver.

\section{Implementation}
We use the public Caffe \cite{jia2014caffe} framework for deep learning. Previous works have shown it is good practice to fine-tune classification networks to segmentation task. We transfer two contemporary  classification models (VGG16 and ResNet). When fine-tuned from the simplified VGG16\footnote{The simplified VGG16 originates from the public available version from DeepLab \url{http://ccvl.stat.ucla.edu/software/deeplab/}. The $4096 \times 7 \times 7 \times 512$ layer and $4096 \times 4096$ layer are sub-sampled to $1024 \times 3 \times 3 \times 512$ and $1024 \times 1024$, which leads to a much smaller model with faster speed.} \cite{simonyan2014very}, weight decay parameter is set to $0.0005$, the momentum parameter is set to $0.99$ and the initial learning rate is set to $10^{-5}$ as we process only one image at each iteration, i.e., mini-batch size is set to 1.

For ResNet, we use our own implementation. We trained the model following the same setting as the authors \cite{he2015deep}. Our own implementation has 56 layers and it gets a  $6.81\%$ top-5 accuracy (standard 10-crops testing) on the ILSRVC 2012 \emph{val} set. The whole training process takes about 10 days on a 4-GPUs architecture. We skip the sub-sampling operation in the $conv5\_1$ layer and modify the filters in the $conv5$ block by introducing zeros to increase the size, which is known as 'hole algorithm' \cite{chen2014semantic}. This operation yields a stride of 16 pixels. Weight decay parameter is set to $0.0001$, the momentum parameter is set to $0.9$ and the initial learning rate is set at $0.01$. The mini-batch size is set to 16, we found the batch size influenced the convergence of the ResNet model, perhaps it is due to the batch normalization layers \cite{ioffe2015batch} . The momentum of batch normalization is set to 0.1, which means that the running mean and variance changes by $10\%$ of its value at each batch.

Scale jittering, color altering \cite{wu2015deep} and horizontal mirror images are adapted for data augmentation. For scale jittering in the training phase, every image is resized with randomly ration in range $[0.5,2.0]$.

\section{Experiments}
\textbf{Datasets.} We test our model on the PASCAL VOC 2012 segmentation benchmark. It includes 20 categories plus background.  The original \emph{train} set has 1464 images with pixel-wise labels. We also use the annotation from~\cite{hariharan2011semantic}, resulting in 10582 (augmented \emph{train} set), 1449 (\emph{val} set) and 1456 (\emph{test} set) images. The accuracy is evaluated by mean IoU scores. Since the \emph{test} set annotation of PASCAL VOC 2012 is not released, the result on \emph{test} set is reported by the server\footnote{\url{http://host.robots.ox.ac.uk:8080}}.

To compare with the-state-of-the-art, we further exploit the large scale dataset MS COCO \cite{lin2014microsoft} to pre-train the model, which includes 123,287 images in its \emph{trainval} set with 80 categories and one background. Each image comes with pixel-wise label.

\begin{figure}
\begin{minipage}[t]{0.5\linewidth}
\centering
\includegraphics[width=1\columnwidth]{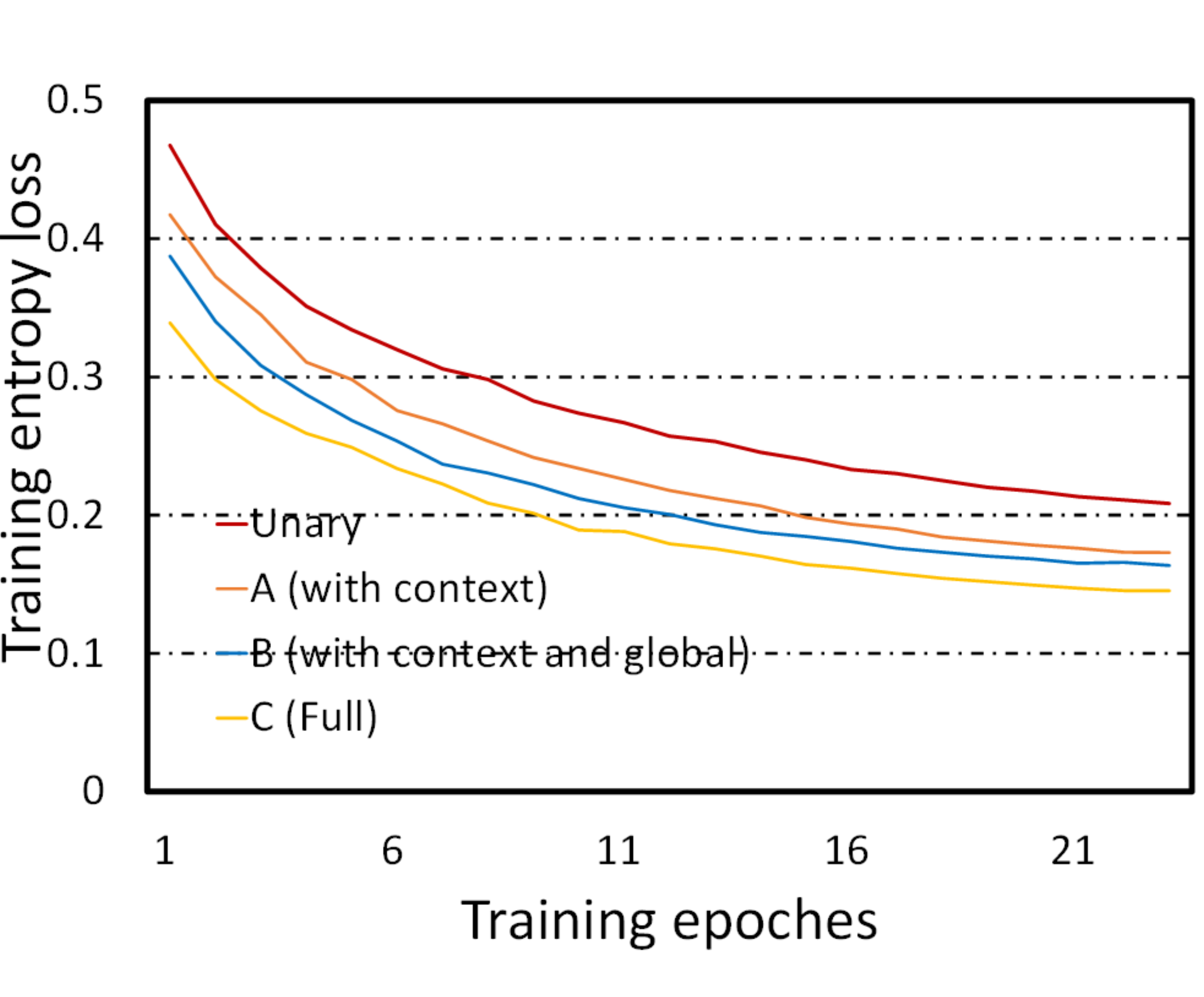}
\captionof{figure}{Training curves.}
\label{figure:traincurves}
\end{minipage}%
\begin{minipage}[t]{0.5\linewidth}
\centering
\vspace{-4.5cm}
\begin{tabular}{l|c}
Component & Time cost (ms) \\
\hline
Unary(VGG) & 44 \\
Unary(ResNet) & 35 \\
High order & 0.7 \\
Global nodes & 0.3 \\
Guidance CRF & 10 \\
\hline
Total(VGG) & 55 \\
Total(ResNet) & 46 \\
\end{tabular}
\vspace{1cm}
\captionof{table}{Inference time for a $500 \times 300$ color image. }
\label{table:time}
\end{minipage}
\end{figure}
\vspace{-10mm}
\subsection{Validation of the Model}
We conduct our evaluations of each components on the Pascal VOC 2012 \emph{val} set (1449 images), training on the augmented \emph{train} set (10582 images), fine-tuning from the simplified VGG16. The detailed settings of \textbf{Unary}, \textbf{A}, \textbf{B} and \textbf{C} are shown in the following paragraph and the performance in Table \ref{table:val} verify the effectiveness of each component in our model.
We train for up to 24 epochs on Pascal VOC 2012 augmented \emph{train} set and the training curves are shown in the Figure \ref{figure:traincurves}, which shows clearly each component of our proposed model can lead to lower training error. The whole training costs about one day for each model on one modern GPU card.

\begin{enumerate}
  \item \textbf{Unary}. We follows the similar settings Deeplab, which is fine-tuned from VGG16 with Pascal VOC 2012 augmented \emph{train} set. We does not adapt bilateral-filtered based fully connected CRF as processing step. This baseline gets a mean IOU of $66.6\%$.
  \item \textbf{A}. The high order term stated in the Equation \ref{equation:highorder} is add to encode context information. We run for only one iteration for computing efficiency. It achieves a mean IOU of $68.2\%$, 1.6 points higher than base-line.  The context information benefits most categories except for ``sofa'' and ``dining table''.
  \item \textbf{B}. This setting uses both  and the global nodes, further boosting the performance by 3.0 points. It verifies the higher order term and the global nodes are helpful to make consistent estimations. Some categories, such as ``sofa'', ``horse'', ``dining table'', have enjoyed great improvements.
  \item \textbf{C}. This setting further adapts the guidance CRF for a sharp boundary and it is the full model. We add the guidance CRF layer and re-train the whole networks from the simplified VGG16 with the same number of training epochs. We run one iteration of mean field in the training stage for computing efficiency and run three iterations for better convergence in the inference stage. The window size of guided filtering is set to 50 by cross validation. It brings the mean IoU score to $73.3\%$.
  \item \textbf{D}. In this setting, we add all components of our proposed model and pre-train the simplified VGG16 model on MS COCO, which is widely adapted by other methods \cite{chen2014learning,zheng2015conditional,lin2015efficient,liu2015semantic}. We get a mean IoU score $74.9\%$ of on the \emph{val} set.
  \item \textbf{E}. We further replace the simplified VGG16 with ResNet. All the other settings follow \textbf{D}. It gets a mean IoU score of $77.6\%$ on the \emph{val} set.
\end{enumerate}

\textbf{Time complexity.} All the code is optimized by CUDA and the time cost is measured on one modern GPU card. For a typical $300 \times 500$ color image, as it is shown in the Table \ref{table:time}, it costs
about $55ms$ in total to compute the segmentation score map on the simplified VGG16.
The whole networks fine-tuned from ResNet process one image within only $46ms$ in total on a modern GPU card, while the unary layers cost $35ms$ and  all the context CRF layers and the guidance CRF layers take about $11ms$, as it is shown in the Table \ref{table:time}.
Our proposed context CRF costs little time while bringing large performance gains.

The bilateral-filtering based fully connected CRF is widely used for sharp object boundary in previous works. The fully connected CRF with a recently optimized implementation of fast bilateral filtering \cite{krahenbuhl2012efficient} takes about $1400ms$ for 10 mean field iterations  on a Intel Xeon(R) CPU W3690. Our proposed guidance CRF costs only $10ms$ on one modern GPU card. We make the process for sharp object boundary more than $100\times$ faster.

\subsection{Comparisons with State-of-art}
 As our model fine-tuned from ResNet performs best on the \emph{val} set, we submit the segmentation result of Pascal VOC 2012 \emph{test} set from this model to the test server. In the test phase, we combine three scales $\{0.8,1.0,1.2\}$ and their horizontal flipped versions to get the predicted score map.

 We quantitatively compare our proposed model with state-of-art models: Deeplab \cite{papandreou2015weakly}, CRF-RNN \cite{zheng2015conditional}, Deeplab-DT \cite{chen2015semantic}, DPN \cite{liu2015semantic} and Piecewise \cite{lin2015efficient},. CRF-RNN and DPN jointly train the filter-based CRF with FCN. Other models adapt bilateral filter based CRF as a post-process step. All these models are trained on the same training data, e.g., ImageNet 2012 \emph{train} set, MS COCO \emph{trainval} set and augmented Pascal VOC 2012 \emph{train} set, for fair comparison.

\begin{table}[H]
\small
\center
\caption{Results on Pascal VOC 2012 \emph{val} set (\%). Unary: on simplified VGG16. A: with context. B: with context and global. C: full. D: with MS COCO. E: on ResNet.}
\vspace{-4mm}
\begin{tabular}{p{2cm}|p{1cm}<{\centering} p{1cm}<{\centering} p{1cm}<{\centering} p{1cm}<{\centering} | p{1cm}<{\centering} p{1cm}<{\centering}}
          & \textbf{Unary}    & \textbf{A}     & \textbf{B}    & \textbf{C}    & \textbf{D}     & \textbf{E}\\
\hline
aeroplane & 80.7    & 83.8  & 81.3 & 84.4  & 84.1  & \textbf{89.3}\\
bicycle    & 33.9    & 36.9  & 37.0 & 37.1  & 37.6  & \textbf{40.8}\\
bird      & 77.4    & 82.0  & 82.6 & 85.4  & 88.8  & \textbf{86.6}\\
boat      & 62.8    & 65.6  & 68.1 & 69.7  & 69.3  & \textbf{70.0}\\
bottle    & 66.0    & 67.0  & 71.4 & 74.0  & 74.4  & \textbf{75.1}\\
bus       & 82.7    & 84.8  & 87.0 & 87.9  & 92.0  & \textbf{94.5} \\
car       & 77.3    & 79.6  & 83.2 & 84.3  & 86.2  & \textbf{88.1}\\
cat       & 81.8    & 85.0  & 86.3 & 88.4  & 90.4  & \textbf{91.4}\\
chair     & 30.5    & 31.4  & 36.2 & 38.3  & 38.5  & \textbf{44.5}\\
cow       & 66.7    & 71.0  & 75.9 & 79.9  & 83.4  & \textbf{87.9}\\
dining table & 52.0 & 43.6  & 56.8 & 57.9  & 59.8  & \textbf{58.9}\\
dog       & 73.4    & 78.2  & 81.3 & 83.7  & 82.5  & \textbf{84.6}\\
horse     & 65.7    & 70.6  & 78.2 & 83.0  & 81.9  & \textbf{90.0}\\
motorbike & 71.9    & 73.1  & 76.6 & 77.0  & 82.2  & \textbf{86.9}\\
person    & 79.5    & 80.1  & 80.5 & 82.2  & 82.5  & \textbf{86.1}\\
potted plant & 46.3  & 49.9  & 48.3 & 52.3  & 54.9  & \textbf{61.2}\\
sheep     & 73.6    & 74.9  & 79.0 & 81.7  & 82.2  & \textbf{86.6}\\
sofa      & 42.8    & 37.1  & 47.7 & 48.5  & \textbf{52.5}  & 52.0\\
train     & 77.7    & 78.1  & 80.5 & 82.6  & 88.3  & \textbf{86.8}\\
tv/monitor & 65.0    & 66.2  & 64.8 & 66.3  & 67.5  & \textbf{75.1}\\
\hline
Mean      & 66.6    & 68.2  & 71.2 & 73.3  & 74.9 & \textbf{77.6}\\
\end{tabular}
\label{table:val}
\end{table}

\vspace{-16mm}

\begin{table}[H]
\small
\center
\caption{Results on Pascal VOC 2012 \emph{test} set (\%).}
\begin{tabular}{p{2cm}|p{1cm}<{\centering} p{1cm}<{\centering} p{1cm}<{\centering} p{1cm}<{\centering} p{1cm}<{\centering} | p{1cm}<{\centering}}
          &\cite{papandreou2015weakly}  & \cite{zheng2015conditional} & \cite{chen2015semantic} & \cite{liu2015semantic} & \cite{lin2015efficient} & \textbf{Ours} \\
\hline
aeroplane & 89.2 & 91.2    & 93.2  & 89.0 & 92.9  & \textbf{93.7} \\
bicycle   & 46.7 & 56.2    & 41.7  & \textbf{61.6} & 39.6  & 39.5 \\
bird      & 88.5 & 88.9    & 88.0  & 87.7 & 84.0  & \textbf{92.9} \\
boat      & 63.5 & 68.0    & 61.7  & 66.8 & 67.9  & \textbf{68.4} \\
bottle    & 68.4 & 70.7    & 74.9  & 74.7 & \textbf{75.3}  & 73.5 \\
bus       & 87.0 & 89.5    & 92.9  & 91.2 & 92.7  & \textbf{94.0} \\
car       & 81.2 & 83.8    & 84.5  & 84.3 & 83.8  & \textbf{85.5} \\
cat       & 86.3 & 87.2    & 90.4  & 87.6 & 90.1  & \textbf{92.8} \\
chair     & 32.6 & 33.6    & 33.0  & 36.5 & \textbf{44.3}  & 36.7 \\
cow       & 80.7 & 81.0    & 82.8  & 86.3 & 85.5  & \textbf{86.8} \\
dining table & 62.4 & 66.4 & 63.2  & 66.1 & 64.9  & \textbf{68.2} \\
dog       & 81.0 & 82.4    & 84.5  & 84.4 & \textbf{87.3}  & 86.5\\
horse     & 81.3 & 83.1    & 85.0  & 87.8 & 88.8  & \textbf{89.7} \\
motorbike & 84.3 & \textbf{87.8}    & 87.2  & 85.6 & 84.5  & 85.9 \\
person    & 82.1 & 82.3    & 85.7  & 85.4 & 85.5  & \textbf{87.6} \\
potted plant & 56.2 & 59.8  & 60.5  & 63.6 & \textbf{68.1}  & 63.7 \\
sheep     & 84.6 & 83.5    & 87.7  & 87.3 & \textbf{89.0}  & 87.2 \\
sofa      & 58.3 & 53.4    & 57.8  & 61.3 & \textbf{62.8}  & 57.2 \\
train     & 76.2 & 79.5    & 84.3  & 79.4 & 81.2  & \textbf{85.4} \\
tv/monitor & 67.2 & 71.1    & 68.2  & 66.4 & \textbf{71.4}  & 70.9 \\
\hline
Mean     & 73.9  & 75.9    & 76.3  & 77.5 & 77.8  & \textbf{78.1}\\
\end{tabular}
\label{table:test}
\end{table}

\begin{figure}[H]
\centering
    \begin{tabular}{c@{~}c@{~}c@{~}|c@{~}c@{~}c@{~}}
\includegraphics[width=0.15\columnwidth]{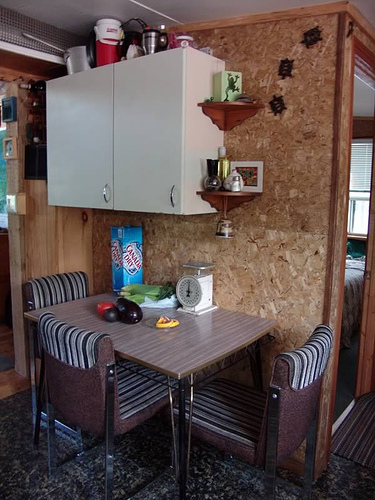}&
\includegraphics[width=0.15\columnwidth]{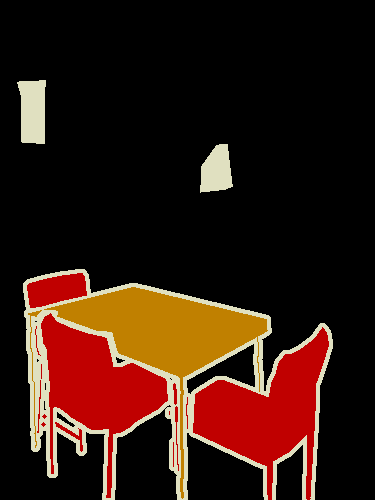}&
\includegraphics[width=0.15\columnwidth]{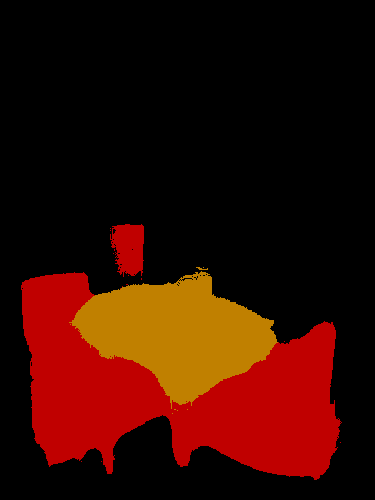}&
\includegraphics[width=0.15\columnwidth]{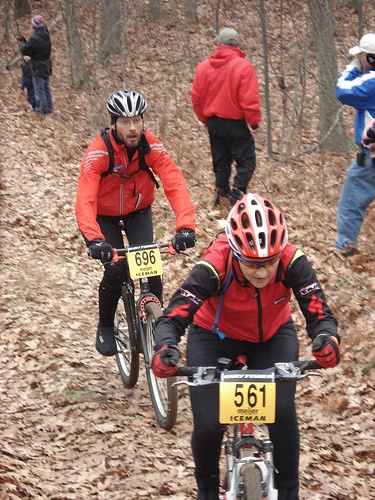}&
\includegraphics[width=0.15\columnwidth]{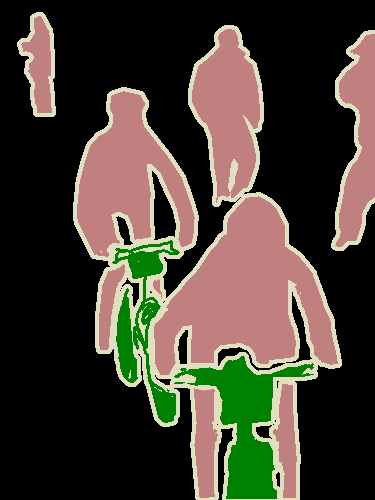}&
\includegraphics[width=0.15\columnwidth]{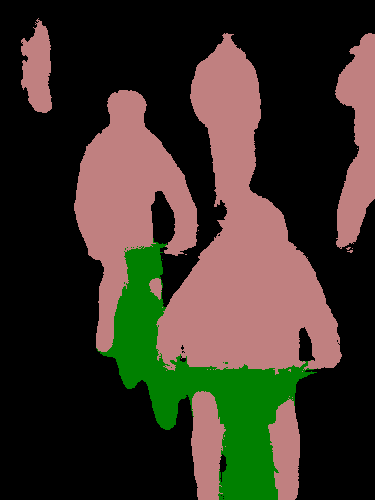}\\

\includegraphics[width=0.15\columnwidth]{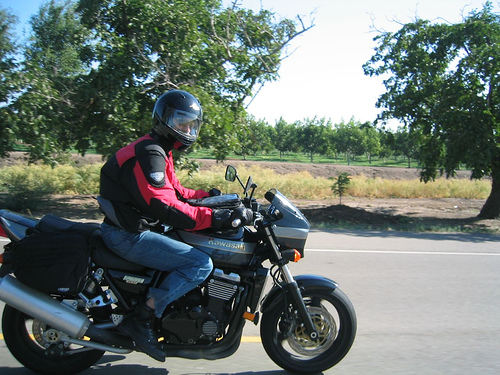}&
\includegraphics[width=0.15\columnwidth]{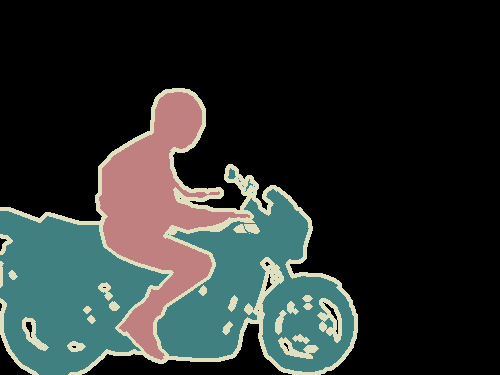}&
\includegraphics[width=0.15\columnwidth]{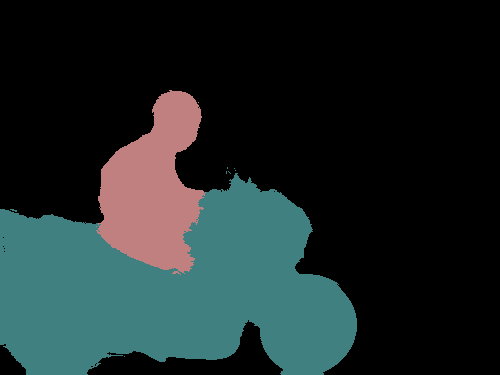}&
\includegraphics[width=0.15\columnwidth]{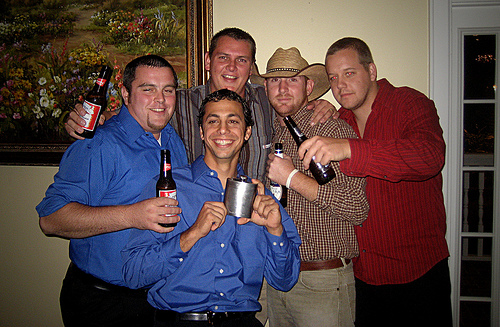}&
\includegraphics[width=0.15\columnwidth]{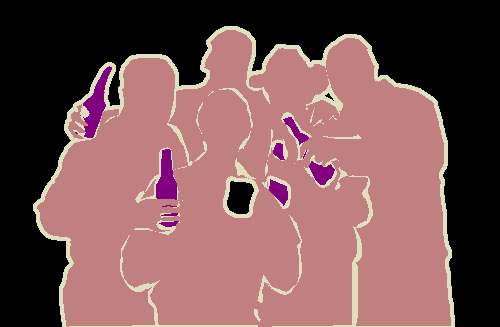}&
\includegraphics[width=0.15\columnwidth]{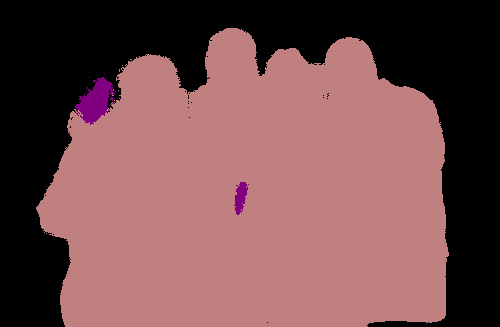}\\

\includegraphics[width=0.15\columnwidth]{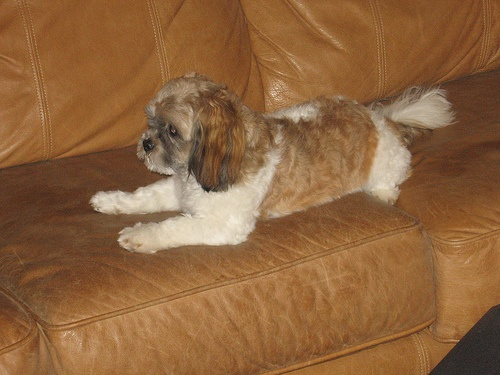}&
\includegraphics[width=0.15\columnwidth]{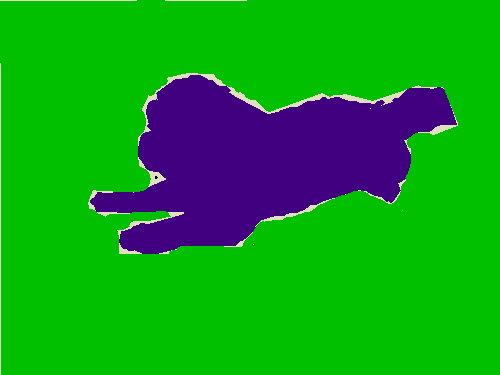}&
\includegraphics[width=0.15\columnwidth]{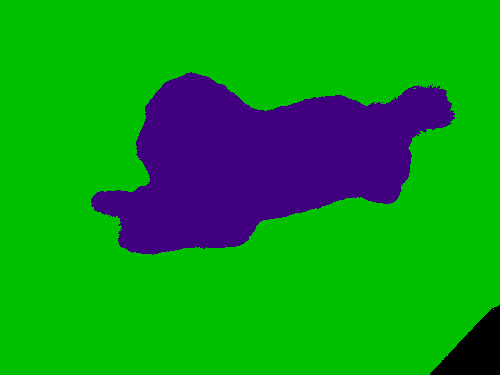}&
\includegraphics[width=0.15\columnwidth]{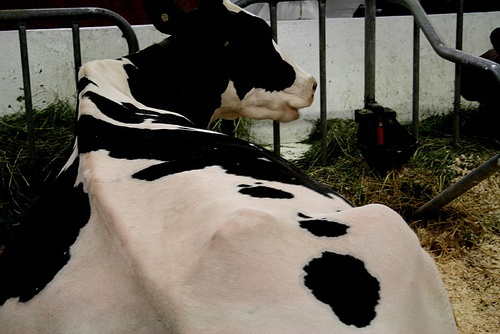}&
\includegraphics[width=0.15\columnwidth]{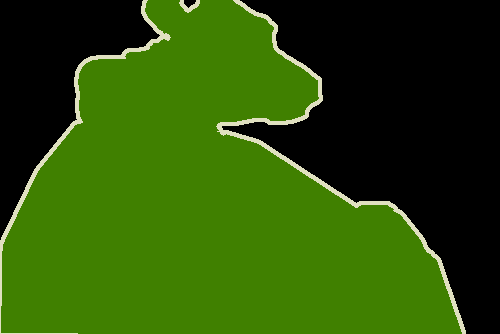}&
\includegraphics[width=0.15\columnwidth]{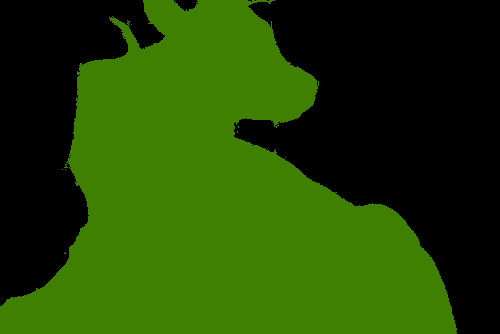}\\

\includegraphics[width=0.15\columnwidth]{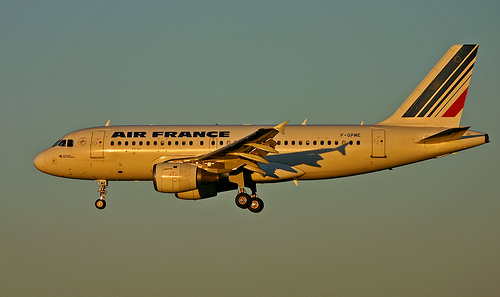}&
\includegraphics[width=0.15\columnwidth]{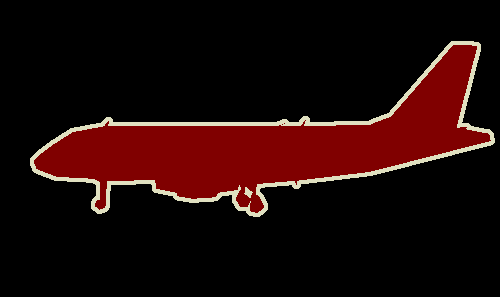}&
\includegraphics[width=0.15\columnwidth]{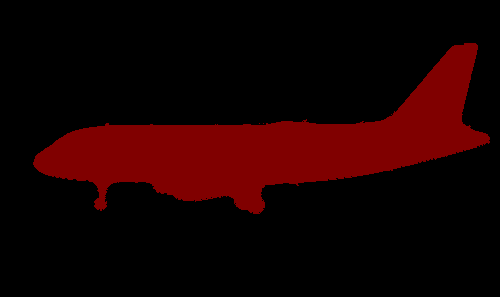}&
\includegraphics[width=0.15\columnwidth]{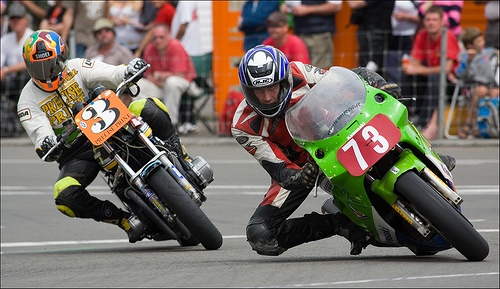}&
\includegraphics[width=0.15\columnwidth]{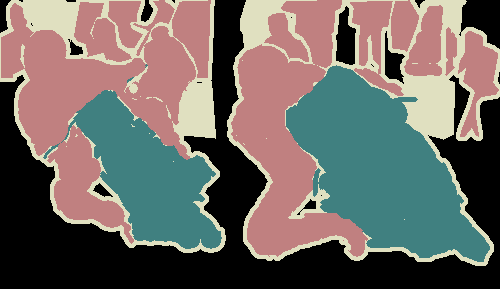}&
\includegraphics[width=0.15\columnwidth]{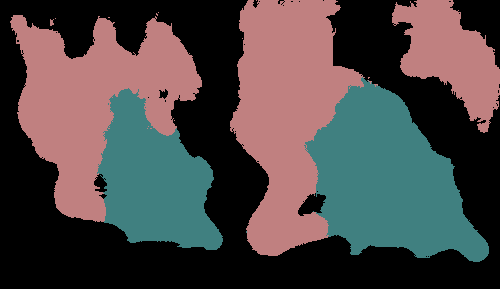}\\

(a) Input & (b) Truth  & (c) Prediction & (a) Input & (b) Truth & (c) Prediction
    \end{tabular}
    \caption{Some examples from Pascal VOC 2012 \emph{val} set.}
    \label{figure:qulative}
\end{figure}

The result of the comparison on Pascal VOC 2012 \emph{test} set is shown in Table \ref{table:test}. We achieve a mean IOU score of $78.1\%$ on this dataset, which outperforms all the existing works. Our model performs the best on more than half of all the 20 categories. Piecewise \cite{lin2015efficient} uses multi-scale feature maps and fine-tunes the model from the complete version of VGG16, which runs much slower than the simplified VGG16. Besides they model lots of pair-wise joint potentials and perform two mean field iterations in the inference stage. They also adapt the bilateral based fully connected CRF for a sharp prediction. As comparison, our model introduce the context CRF and the guidance CRF to encode context information and delineate the object boundary, which run much faster with higher performance on the overall performance. Some examples are shown in the Figure \ref{figure:qulative}.

\vspace{-4mm}
\section{Conclusion}
\vspace{-2mm}
In this paper, we have proposed a deep coarse-to-fine model with high order context and guided filtering for semantic image segmentation.
At the coarse level, we target directly at the influence of high order pattern on the unary node to encode the relative interaction between them. We also introduce the hidden global nodes to keep global predictions and local predictions consistent.
At the fine level, instead of bilateral filtering based CRF, we plug in the guided filtering as one step of message passing in the mean field algorithm and make it $100\times$ faster to delineate the object boundary.
We transfer two contemporary image classification models for the task of semantic image segmentation. Experiments on the Pascal VOC 2012 dataset show that our model outperforms the state-of-the-art performance with appealing running speed, which demonstrates our model can harness context information effectively for structural prediction and can locate the object accurately.
\vspace{-5mm}

\bibliographystyle{splncs}
\bibliography{mybib}

\end{document}